\documentclass[runningheads]{llncs}
\usepackage[T1]{fontenc}
\usepackage{graphicx}
\usepackage{amsmath,amssymb}
\usepackage{mathtools}
\usepackage{hyperref}
\usepackage[capitalize,noabbrev,nameinlink]{cleveref}
\usepackage[dvipsnames]{xcolor}
\usepackage{booktabs}
\usepackage{multirow}
\usepackage{etoolbox}
\usepackage[]{caption}
\usepackage{subcaption}
\usepackage{float}
\let\llncssubparagraph\subparagraph
\let\subparagraph\paragraph
\usepackage{titlesec}
\let\subparagraph\llncssubparagraph
\usepackage{enumitem}
\usepackage{stmaryrd}
\usepackage{apptools}
\usepackage{textcomp}
\usepackage{nicefrac}
\usepackage{xfrac}

\usepackage{pgf}
\usepackage{tikz}
\usepackage{forest}

\usetikzlibrary{arrows,decorations.pathmorphing,decorations.pathreplacing,decorations.footprints,fadings,calc,trees,mindmap,shadows,decorations.text,patterns,positioning,shapes,matrix,fit}
\usetikzlibrary{arrows,shadows,backgrounds}
\usetikzlibrary{arrows.meta}
\usetikzlibrary{positioning,fit}
\usetikzlibrary{automata}
\usetikzlibrary{matrix}
\usetikzlibrary{shapes.symbols,shapes.misc,shapes.arrows}
\usetikzlibrary{matrix,chains,scopes,decorations.pathmorphing}
\usetikzlibrary{bayesnet}
\usetikzlibrary{tikzmark}

\definecolor{darkred}{rgb}{0.7,0.1,0.1}
\definecolor{medred}{rgb}{0.5,0.1,0.1}
\definecolor{midred}{rgb}{0.7,0.2,0.2}
\definecolor{vdarkred}{rgb}{0.4,0.1,0.1}
\definecolor{darkslategray}{rgb}{0.18, 0.31, 0.31} 
\definecolor{platinum}{rgb}{0.9, 0.89, 0.89} 
\definecolor{gray}{rgb}{.4,.4,.4}
\definecolor{midgrey}{rgb}{0.5,0.5,0.5}
\definecolor{middarkgrey}{rgb}{0.35,0.35,0.35}
\definecolor{darkgrey}{rgb}{0.3,0.3,0.3}
\definecolor{darkred}{rgb}{0.7,0.1,0.1}
\definecolor{midblue}{rgb}{0.2,0.2,0.7}
\definecolor{darkblue}{rgb}{0.1,0.1,0.5}
\definecolor{darkgreen}{rgb}{0.1,0.5,0.1}
\definecolor{defseagreen}{cmyk}{0.69,0,0.50,0}
\definecolor{purple3}{RGB}{125,38,205}          

\definecolor{tyellow1}{HTML}{FCE94F}
\definecolor{tyellow2}{HTML}{EDD400}
\definecolor{tyellow3}{HTML}{C4A000}
\definecolor{torange1}{HTML}{FCAF3E}
\definecolor{torange2}{HTML}{F57900}
\definecolor{torange3}{HTML}{C35C00}
\definecolor{tbrown1}{HTML}{E9B96E}
\definecolor{tbrown2}{HTML}{C17D11}
\definecolor{tbrown3}{HTML}{8F5902}
\definecolor{tgreen1}{HTML}{8AE234}
\definecolor{tgreen2}{HTML}{73D216}
\definecolor{tgreen3}{HTML}{4E9A06}
\definecolor{tblue1}{HTML}{729FCF}
\definecolor{tblue2}{HTML}{3465A4}
\definecolor{tblue3}{HTML}{204A87}
\definecolor{tpurple1}{HTML}{AD7FA8}
\definecolor{tpurple2}{HTML}{75507B}
\definecolor{tpurple3}{HTML}{5C3566}
\definecolor{tred1}{HTML}{EF2929}
\definecolor{tred2}{HTML}{CC0000}
\definecolor{tred3}{HTML}{A40000}
\definecolor{tlgray1}{HTML}{EEEEEC}
\definecolor{tlgray2}{HTML}{D3D7CF}
\definecolor{tlgray3}{HTML}{BABDB6}
\definecolor{tdgray1}{HTML}{888A85}
\definecolor{tdgray2}{HTML}{555753}
\definecolor{tdgray3}{HTML}{2E3436}

\newcommand{\dghlight}[1]{{\color[RGB]{0,120,0}#1}}

\newcommand{\ncolor}[1]{{\color{tbrown3}#1}}

\patchcmd{\paragraph}{\itshape}{\bfseries\boldmath}{}{}

\hypersetup{
  colorlinks = true,
  linkcolor = MidnightBlue,
  urlcolor  = MidnightBlue,
  citecolor = MidnightBlue,
  anchorcolor = MidnightBlue
}

\definecolor{gray}{rgb}{.4,.4,.4}
\definecolor{midgrey}{rgb}{0.5,0.5,0.5}
\definecolor{middarkgrey}{rgb}{0.35,0.35,0.35}
\definecolor{darkgrey}{rgb}{0.3,0.3,0.3}
\definecolor{darkred}{rgb}{0.7,0.1,0.1}
\definecolor{midblue}{rgb}{0.2,0.2,0.7}
\definecolor{darkblue}{rgb}{0.1,0.1,0.5}
\definecolor{defseagreen}{cmyk}{0.69,0,0.50,0}

\newcommand{\fml}[1]{{\mathcal{#1}}}

\newcommand{\tn}[1]{\textnormal{#1}}

\newcommand{\msf}[1]{\ensuremath\mathsf{#1}}
\newcommand{\mbf}[1]{\ensuremath\mathbf{#1}}
\newcommand{\mbb}[1]{\ensuremath\mathbb{#1}}
\newcommand{\oper}[1]{\ensuremath\mathsf{#1}}
\newcommand{\vars}{\oper{vars}}
\newcommand{\nfrac}{\nicefrac}
\newcommand{\cf}{\ensuremath\upsilon} 
\newcommand{\cfn}[1]{\ensuremath\upsilon_{#1}} 

\newcommand{\shapname}{Sv}
\newcommand{\svn}[1]{\msf{\shapname}_{#1}}

\newcommand{\sv}{\msf{\shapname}}

\newcommand{\waxp}{\ensuremath\msf{WAXp}}
\newcommand{\wcxp}{\ensuremath\msf{WCXp}}
\newcommand{\axp}{\ensuremath\msf{AXp}}
\newcommand{\cxp}{\ensuremath\msf{CXp}}

\newcommand{\exv}{\ensuremath\mathbf{E}}

\newcommand{\irrelevant}{\oper{Irrelevant}}

\DeclareMathOperator*{\entails}{\vDash}

\DeclareMathOperator*{\limply}{\rightarrow}

\makeatletter
\newcommand\nparagraph{%
  \@startsection{paragraph}
    {4}
    {\z@}
    {1.5ex \@plus1ex \@minus.2ex}
    {-1em}
    {\normalfont\normalsize\bfseries}%
}
\makeatother

\renewcommand{\paragraph}{\nparagraph}
\newcommand{\similar}{\ensuremath\sigma}
\newcommand{\tsimilar}{\ensuremath\mathsf{T}\sigma}

\newcommand{\mdist}{\ensuremath\mathfrak{d}}

\makeatletter
\AddToHook{cmd/appendix/before}{\def\cref@section@alias{appendix}\def\cref@subsection@alias{appendix}}
\makeatother

\tikzset{
  0 my edge/.style={densely dashed, my edge},
  my edge/.style={-{Stealth[]}},
}

\begin{document}
\title{Towards Rigorous Explainability\\by Feature Attribution}
%
\titlerunning{Rigorous Explainability by Feature Attribution}
%
\author{%
  Olivier~L\'{e}toff\'{e}\inst{1}\orcidID{ 0009-0001-0035-0444}
  \and
  Xuanxiang~Huang\inst{2}\orcidID{0000-0002-3722-7191}
  \and
  Joao~Marques-Silva\inst{3}\orcidID{0000-0002-6632-3086}
}
\authorrunning{O. L\'{e}toff\'{e} et al.}
%
\institute{
  IRIT, University of Toulouse France \\
  \email{olivier.letoffe@orange.fr}\\
  \and
  Nanyang Technological University, Singapore\\
  \email{xuanxiang.huang@ntu.edu.sg}
  \and
  ICREA \& Univ.\ Lleida, Spain\\
  \email{jpms@icrea.cat}
}
\let\oldaddcontentsline\addcontentsline
\def\addcontentsline#1#2#3{}
\maketitle              
\def\addcontentsline#1#2#3{\oldaddcontentsline{#1}{#2}{#3}}
\begin{abstract}
  For around a decade, non-symbolic methods have been the option of
  choice when explaining complex machine learning (ML) models.
  Unfortunately, such methods lack rigor and can mislead human 
  decision-makers.
  In high-stakes uses of ML, the lack of rigor is especially
  problematic.
  One prime example of provable lack of rigor is the adoption of
  Shapley values in explainable artificial intelligence (XAI), with
  the tool SHAP being a ubiquitous example. This paper overviews 
  the ongoing efforts towards using rigorous symbolic methods of XAI
  as an alternative to non-rigorous non-symbolic approaches,
  concretely for assigning relative feature importance.
\end{abstract}

%


\section{Overview}

The remarkable advances in machine learning (ML) are not without
shortcomings.
Among these, lack of interpretability is paramount.
Motivated by the complexity of ML models, human decision makers are
often unable to understand the rationale for the predictions made.
Thus, one key goal of eXplainable Artificial Intelligence (XAI) is to
help human decision makers in fathoming the predictions of ML models.

The explanation of ML models can be broadly organized into
explanations by feature selection and those by feature
attribution~\cite{molnar-bk20}.
Explanations by feature selection can be interpreted as answering a
\emph{Why?} question, i.e.\ if the selected features take their
assigned values, then the prediction is the one we want to explain.
In contrast, feature attribution aims to assign a value to each
feature, representing the relative importance of the feature for the
prediction.

The best-known method of explainability by feature attribution is the
tool SHAP~\cite{lundberg-nips17}, which finds ubiquitous uses in ML.
SHAP is based on the well-known Shapley values from game
theory~\cite{shapley-ctg53}, and computes the so-called SHAP scores.
A cooperative game is a pair $G=(N,\upsilon)$, where $N$ is a set of
players, e.g.\ the players in a game, the voters in an election, or
the features in an ML model, and $\upsilon$ is a characteristic
function that maps sets of features to the 
real values, $\upsilon:2^{N}\to\mathbb{R}$.
Given a characteristic function, the Shapley values are unique.
Furthermore, for each characteristic function a different set of
Shapley values is obtained.

This paper provides a brief overview of the recent efforts that
demonstrated critical flaws in the most often used definition of
Shapley values for XAI, i.e.\ the definition used in
SHAP~\cite{msh-cacm24,hms-ijar24}.
As detailed in more recent work, the SHAP scores can produce
misleading information because of the characteristic function that has
been used in
XAI~\cite{kononenko-jmlr10,kononenko-kis14,lundberg-nips17}.
In addition, this paper outlines more recent work on finding
alternatives to the widely used (but flawed) SHAP
scores~\cite{lhms-aaai25,mshl-corr25}.
This recent work involves proposing a new characteristic function, and
devising novel algorithms aiming at the efficient computation of
corrected SHAP scores.
The novel characteristic function is based on a logic definition of
explanation by feature selection~\cite{ms-rw22}, and so it establishes
a link between two well-known kinds of explanation, i.e.\ feature
selection and feature attribution, but also between Shapley values and
logic-based definitions of explanations.
At a more abstract level, this work also uncovers new links between
formal methods and game theory.

The paper is organized as follows.
\cref{sec:prelim} introduces the notation and definitions used in the
remainder of the paper, and
\cref{sec:fxai} completes the preliminaries by providing a brief
glimpse of logic-based XAI.
Afterwards, \cref{sec:svmyth} overviews the known flaws of SHAP
scores.
The remainder of the paper summarizes our recent work towards
developing a trustable alternative to SHAP scores.
\cref{sec:fixsv} surveys a recent proposal of SHAP scores that
corrects the known flaws of past definitions. These will be referred
to as \emph{corrected SHAP scores}.
Then, \cref{sec:nushap} details practical methods for computing the
corrected SHAP scores.
\cref{sec:res} provides a glimpse of existing experimental results,
and \cref{sec:conc} concludes the paper.

\section{Preliminaries} \label{sec:prelim}

\paragraph{Machine learning (ML) models.}
%
An ML model is defined on a set $\fml{F}=\{1,\ldots,m\}$ of features.
Each feature $i\in\fml{F}$ takes values from a domain $\mbb{D}_i$.
Domains can be categorical, integer- or real-valued.
The feature space $\mbb{F}$ is defined as the cartesian product (in
order) of the domains of the features,
i.e.\ $\mbb{F}=\mbb{D}_1\times\cdots\times\mbb{D}_m$.
Given a set $\mbb{V}$ of values, the ML model computes a non-constant
prediction function $\pi:\mbb{F}\to\mbb{V}$.
An ML model $M$ is defined as a tuple
$M=(\fml{F},\mbb{F},\mbb{V},\pi)$.
Observe that the definition of ML model applies both to classification
and regression models.
For classification problems, $\mbb{V}$ is a finite set of values,
which may be categorical. For regression problems, $\mbb{V}$ may be
infinite, and there are no restriction on the values  of $\mbb{V}$.
An instance $I$ is a pair $I=(\mbf{v},p)$ where $\mbf{v}\in\mbb{F}$
and $p=\pi(\mbf{v})$.

An explanation problem is a tuple $\fml{E}=(M,I)$, where $M$ can
either be a classification or a regression model, and $I=(\mbf{v},p)$
is a given instance, with $\mbf{v}\in\mbb{F}$.
(Observe that $p=\pi(\mbf{v})$, with $p\in\mbb{V}$.)

\paragraph{Running examples.}
%
\cref{fig:ex01} shows one of the running examples used throughout the
paper. This example was first studied in~\cite{msh-cacm24}.

\begin{figure*}[t]
  \begin{subfigure}[b]{0.3125\textwidth}
    \scalebox{0.875}{
      \renewcommand{\tabcolsep}{0.45em}
\begin{tabular}{ccccc} \toprule
  row$\,$\# & $x_1$ & $x_2$ & $x_3$ & $\pi_1(\mbf{x})$ 
  \\ \toprule
  1 & 0 & 0 & 0 & 0 \\
  2 & 0 & 0 & 1 & 4 \\
  3 & 0 & 0 & 2 & 0 \\
  4 & 0 & 1 & 0 & 0 \\
  5 & 0 & 1 & 1 & 7 \\
  6 & 0 & 1 & 2 & 0 \\
  %
  7 & 1 & 0 & 0 & 1 \\
  8 & 1 & 0 & 1 & 1 \\
  9 & 1 & 0 & 2 & 1 \\
  10 & 1 & 1 & 0 & 1 \\
  11 & 1 & 1 & 1 & 1 \\
  \tikzmarknode{a}{12} & 1 & 1 & 2 & \tikzmarknode{b}{1} 
  \\ \bottomrule
  \begin{tikzpicture}[overlay,remember picture]
    \node[draw=midblue, thin, xshift=-0.35pt, yshift=-0.35pt, inner
      sep=2.0pt, fit=(a) (b)] {};
  \end{tikzpicture}
\end{tabular}

    }
    \caption{Tabular representation} \label{ex01:tr}
  \end{subfigure}
  \begin{subfigure}[b]{0.345\textwidth}
    \scalebox{0.875}{
%
\forestset{
  BDT/.style={
    for tree={
      l=1.5cm,s sep=1.15cm,
      if n children=0{}{circle}, 
      draw=midblue,
      text=midblue,
      edge={
        my edge
      },
      edge=thick,
    }
  },
}
\begin{forest}
  BDT
  [{$x_1$}, label={[yshift=-7.0ex]{{\tiny1}}} 
    [{$x_3$}, label={[yshift=-7.0ex]{{\tiny2}}}, 
      edge label={node[midway,left,xshift=-0.5pt] {{\scriptsize$\in\{0\}$}}}
      [{$x_2$}, label={[yshift=-7.0ex]{{\tiny4}}}, 
        edge label={node[midway,left,xshift=-1.5pt] {{\scriptsize$\in\{1\}$}}}
        [\dghlight{\textbf{4}}, label={[yshift=-5.25ex]{{\tiny6}}},
          edge label={node[midway,left,xshift=-0.5pt] {{\scriptsize$\in\{0\}$}}}, rectangle, fill={tblue2!25} ]
        [\dghlight{\textbf{7}}, label={[yshift=-5.25ex]{{\tiny7}}},
          edge label={node[midway,right,xshift=-0.575pt] {{\scriptsize$\in\{1\}$}}}, rectangle, fill={tblue2!25} ]
      ]
      [\dghlight{\textbf{0}}, label={[yshift=-5.25ex]{{\tiny5}}},
        edge label={node[midway,right,xshift=-0.5pt] {{\scriptsize$\in\{0,2\}$}}},
        rectangle, fill={tblue2!20} ]
    ]
    [\dghlight{\textbf{1}}, label={[yshift=-5.25ex]{{\tiny3}}},
      edge={very thick,draw=purple3}, edge label={node[midway,right,xshift=0.5pt] {{\scriptsize$\in\{1\}$}}},
      rectangle, fill={tblue2!25} ]
  ]
\end{forest}
    }
    \\[-8pt]
    \caption{Decision tree} \label{ex01:dt}
  \end{subfigure}
  \begin{subfigure}[b]{0.3175\textwidth}
    \scalebox{0.875}{
      \renewcommand{\tabcolsep}{0.45em}
\begin{tabular}{ccc} \toprule
  $\fml{S}$ & $\msf{rows}(\fml{S})$ & $\cfn{e}(\fml{S})$
  \\ \toprule
  $\emptyset$ & $1..12$ & $\nfrac{17}{12}$ \\
  $\{1\}$ & $7..12$ & $1$ \\
  $\{2\}$ & $4..6,10..12$ & $\nfrac{10}{6}$ \\
  $\{3\}$ & $3,6,9,12$ & $\nfrac{1}{2}$ \\
  $\{1,2\}$ & $10..12$ & $1$ \\
  $\{1,3\}$ & $9,12$ & $1$ \\
  $\{2,3\}$ & $6,12$ & $\nfrac{1}{2}$ \\
  $\{1,2,3\}$ & $12$ & $1$ 
  \\ \bottomrule
\end{tabular}

    }
    ~\\[-2pt]
    \caption{Expected values} \label{ex01:avg}
  \end{subfigure}
  \caption{Classification model $M_1$ represented as a decision tree.}
  \label{fig:ex01}
\end{figure*}

\begin{example}
  For the ML model shown in~\cref{fig:ex01}, we have
  $\fml{F}_1=\{1,2,3\}$, $\mbb{D}_{1,1}=\mbb{D}_{1,2}=\{0,1\}$,
  $\mbb{D}_{1,3}=\{0,1,2\}$,
  $\mbb{F}_1=\mbb{D}_{1,1}\times\mbb{D}_{1,2}\times\mbb{D}_{1,3}$,
  and
  $\mbb{V}_1=\{0,1,4,7\}$. The ML model $M_1$ is represented by the
  tuple $M_1=(\fml{F}_1,\mbb{F}_1,\mbb{V}_1,\pi_1)$, where $\pi_1$ is 
  shown in~\cref{ex01:tr,ex01:dt}, respectively as a tabular
  representation (TR) and as a decision tree (DT). The target instance
  is $I_1=((1,1,2),1)$ as shown. The explanation problem is thus
  $\fml{E}_1=(M_1,I_1)$.
\end{example}

\cref{fig:ex02} shows the second running example that is used
throughout the paper. This example was first studied
in~\cite{lhms-aaai25}, and represents a regression model with discrete
domains and discrete predicted values.

\begin{figure*}[t]
  \begin{subfigure}[b]{0.35\linewidth} 

    \begin{tabular}{c} ~~\\[5pt] \end{tabular}

    \centering
    \scalebox{0.875}{
      \renewcommand{\arraystretch}{1.15}
      \renewcommand{\tabcolsep}{0.575em}
      \renewcommand{\tabcolsep}{0.45em}
\begin{tabular}{cccc} \toprule
  row$\,$\# & $x_1$ & $x_2$ & $\pi_2(\mbf{x})$ 
  \\ \toprule
  1 & 0 & 0 & $\nfrac{-1}{2}$ \\
  2 & 0 & 1 & $\nfrac{3}{2}$ \\
  3 & 1 & 0 & 1 \\
  \tikzmarknode{a}{4} & 1 & 1               & \tikzmarknode{b}{1}
  \\ \bottomrule
  \begin{tikzpicture}[overlay,remember picture]
    \node[draw=midblue, thin, xshift=-0.35pt, yshift=-0.35pt, inner
      sep=2.0pt, fit=(a) (b)] {};
  \end{tikzpicture}
\end{tabular}

    }
    %
    %
    \caption{Tabular representation} 
    \label{ex02:tr}
  \end{subfigure}
  \begin{subfigure}[b]{0.3\linewidth} 
    \centering
    \scalebox{0.85}{
%
\forestset{
  BDT/.style={
    for tree={
      l=1.5cm,s sep=1.15cm,
      if n children=0{}{circle}, 
      draw=midblue,
      text=midblue,
      edge={
        my edge
      },
      edge=thick,
    }
  },
}
\begin{forest}
  BDT
  [{$x_1$}, label={[yshift=-7.0ex]{{\tiny1}}} 
    [{$x_2$}, label={[yshift=-7.0ex]{{\tiny2}}}, 
      edge label={node[midway,left,xshift=-1.5pt] {{\scriptsize$\in\{0\}$}}}
      [\ncolor{\boldmath{$\nfrac{-1}{2}$}}, label={[yshift=-6.0ex]{{\tiny4}}},
        edge label={node[midway,left,xshift=-0.5pt]
          {{\scriptsize$\in\{0\}$}}}, rectangle, fill={torange1!10} ]
      [\ncolor{\boldmath{$\nfrac{3}{2}$}}, label={[yshift=-6.0ex]{{\tiny5}}},
        edge label={node[midway,right,xshift=-0.575pt] {{\scriptsize$\in\{1\}$}}}, rectangle, fill={torange1!10} ]
    ]
    [\ncolor{\textbf{1}}, label={[yshift=-5.25ex]{{\tiny3}}},
      edge={very thick,draw=tblue2}, edge label={node[midway,right,xshift=0.5pt] {{\scriptsize$\in\{1\}$}}},
      rectangle, fill={torange1!10} ]
  ]
\end{forest}
    }
    \caption{Regression tree (RT)}
    \label{ex02:rt}
  \end{subfigure}
  \begin{subfigure}[b]{0.325\linewidth} 

    \begin{tabular}{c} ~~\\[5pt] \end{tabular}

    \centering
    \scalebox{0.875}{
      \renewcommand{\arraystretch}{1.15}
      \renewcommand{\tabcolsep}{0.575em}
      \renewcommand{\tabcolsep}{0.45em}
\begin{tabular}{ccc} \toprule
  $\fml{S}$ & $\msf{rows}(\fml{S})$ & $\cfn{e}(\fml{S})$
  \\ \toprule
  $\emptyset$ & $1,2,3,4$ & $\nfrac{3}{4}$ \\
  $\{1\}$ & $3,4$ & $1$ \\
  $\{2\}$ & $2,4$ & $\nfrac{5}{4}$ \\
  $\{1,2\}$ & $4$ & $1$
  \\ \bottomrule
\end{tabular}

    }

    \begin{tabular}{c} ~~\\[-1.0pt] \end{tabular}

    \caption{Expected values} \label{ex02:avg}
  \end{subfigure}
  \caption{Regression model $M_2$ represented as a regression tree.}
  \label{fig:ex02}
\end{figure*}

\begin{example}
  For the ML model shown in~\cref{fig:ex02}, we have
  $\fml{F}_2=\{1,2\}$, $\mbb{D}_{2,1}=\mbb{D}_{2,2}=\{0,1\}$,
  $\mbb{F}_2=\mbb{D}_{2,1}\times\mbb{D}_{2,2}$,
  and
  $\mbb{V}_1=\{1,-\nfrac{1}{2},\nfrac{3}{2}\}$. The ML model $M_2$ is
  represented by the tuple
  $M_2=(\fml{F}_2,\mbb{F}_2,\mbb{V}_1,\pi_2)$, where $\pi_2$ is  
  shown in~\cref{ex02:tr,ex02:rt}, respectively as a tabular
  representation (TR) and as a regression tree (DT). The target
  instance is $I_2=((1,1),1)$ as shown. The explanation problem is thus
  $\fml{E}_2=(M_2,I_2)$.
\end{example}

The expected values shown in both~\cref{ex01:avg,ex02:avg} are
discussed later in the paper. The definition of expected values is
standard, e.g.~\cite{lhms-aaai25}.

\paragraph{Lipschitz continuity.}
%
Let $(\mbb{F},\mdist_F)$ and $(\mbb{V},\mdist_V)$ denote metric
spaces.%
\footnote{%
$\mdist_F:\mbb{F}\times\mbb{F}\to\mbb{R}$
and 
$\mdist_V:\mbb{V}\times\mbb{V}\to\mbb{R}$
denote distance functions between two points, which we refer to as
$\mdist$. A distance function $\mdist$ respects the well-known axioms:  
(i) $\mdist(\mbf{x},\mbf{x})=0$;
(ii) if $\mbf{x}\not=\mbf{y}$, then $\mdist(\mbf{x},\mbf{y})>0$;
(iii) $\mdist(\mbf{x},\mbf{y})=\mdist(\mbf{y},\mbf{x})$; and
(iv)
$\mdist(\mbf{x},\mbf{z})\le\mdist(\mbf{x},\mbf{y})+\mdist(\mbf{y},\mbf{z})$.
Examples of distance functions include Hamming, Manhattan, Euclidean
and other distance defined by norm $l_p$, $p\ge1$,
where
$\lVert\mbf{x}\rVert_{p}:=\left(\sum\nolimits_{i=1}^{m}|x_i|^{p}\right)^{\sfrac{1}{p}}$.
%
}
A regression function $\pi:\mbb{F}\to\mbb{V}$ is
Lipschitz-continuous~\cite{osearcoid-bk06} if there exists a constant
$C\ge0$ such that,
\begin{equation} \label{def:lipc}
  \forall(\mbf{x}_1,\mbf{x}_2\in\mbb{F}).
  \mdist_V(\pi(\mbf{x}_1),\pi(\mbf{x}_2))\le{C}\mdist_F(\mbf{x}_1,\mbf{x}_2),
\end{equation}
where $C$ is referred to as the Lipschitz constant.
It is well-known that any Lipschitz-continuous function is also
continuous.
The relationship between Lipschitz continuity and adversarial
robustness has been acknowledged for more than a
decade~\cite{szegedy-iclr14}, i.e.\ since the brittleness of ML models
was recognized as a significant limitation of neural networks
(NNs).

\paragraph{Additional notation.}
%
Given an instance $(\mbf{v},p)$, the notation
$\mbf{x}_{S}=\mbf{v}_{S}$ is defined by, 
\begin{equation}
  \mbf{x}_{S}=\mbf{v}_{S}\quad\coloneq\quad\left(\bigwedge\nolimits_{i\in{S}}x_i=v_i\right)
\end{equation}
Thus, given $S\in\fml{F}$, 
\[
\Upsilon(S)\quad \coloneq \quad\{\mbf{x}\in\mbb{F}\,|\,\mbf{x}_{S}=\mbf{v}_{S}\}
\]
denotes the set of points $\mbf{x}$ in feature space such that the
features represented by $S$ take the values specified by the instance
$(\mbf{v},p)$.

\paragraph{Shapley values.}
%
Following~\cite{elkind-bk12}, we define a 
game $G$ as a pair $(N,\nu)$, 
where $N$ is a finite non-empty set
and $\nu:2^{N}\to\mbb{R}$ is a characteristic function.
The elements of $N$ are referred to as the players, the voters, but
also as the features.
The characteristic function assigns a value to each coalition,
i.e.\ subset of the players.
%
%
In the context of XAI, $N$ corresponds to the set of features
$\fml{F}$.

The Shapley value represents one way of dividing the worth of a game
(i.e.\ $\cf(N)$) by its players, one that respects a number of
important properties. 
For each element $i\in{N}$, given a characteristic function $\cf_{t}$,
its Shapley value is defined by:
\begin{equation}
  \sv_t(i) =
  \sum\nolimits_{S\subseteq{N\setminus\{i\}}}\varsigma(S)\times\Delta_i(S)
\end{equation}
where,
\begin{align}
  \Delta_i(S) & = (\cf_t(S\cup\{i\})-\cf_t(S))\nonumber \\[3.5pt]
  \varsigma(S) & = \nicefrac{|S|!(|N|-|S|-1)!}{|N|!}\nonumber 
\end{align}
The complexity of computing the Shapley values is unwieldy. For
example, for weighted voting games~\cite[Chapter~4]{elkind-bk12}, it
is known that decision problem of computing the Shapley values is
\#P-complete~\cite[Theorem~9]{papadimitriou-mor94}.
As a result, approximation algorithms have often been
proposed~\cite{tejada-cor09}.
Furthermore, alternatives to the Shapley value include the Banzhaf
index~\cite{banzhaf-rlr65}, among many others. These alternatives are
often studied in the context of measuring a priori voting
power~\cite{machover-hscv15}, one example being weighted voting games.

\paragraph{SHAP scores.}
%
The original work on SHAP~\cite{lundberg-nips17} adopted a concrete
game for XAI and an algorithm for approximating the computation of the
Shapley value for that game.
The game for XAI builds on earlier
work~\cite{kononenko-jmlr10,kononenko-kis14}, where $N$ denotes the
set of features $\fml{F}$, and the characteristic function is given
by,
\begin{equation} \label{eq:defsve}
  \cfn{e}(S)=\mbf{E}[\pi(\mbf{x})\,|\,\mbf{x}_{S}=\mbf{v}_{S}]
\end{equation}
i.e.\ for the XAI game, the characteristic function is defined as the
expected value of the classifier when the features in $S$ are fixed to
the values dictated by $\mbf{v}$. For each feature $i\in\fml{F}$, the
SHAP score for $i$ obtained with~\eqref{eq:defsve} is denoted by
$\svn{e}(i)$.%
\footnote{%
The computation of expected values depends on the type of features
used. The definitions proposed in~\cite{lhms-aaai25} are assumed.}
Furthermore, given the game adopted for XAI, the tool SHAP uses a
dedicated method to estimate the expected value for a given set
$S$~\cite{lundberg-nips17}.


\begin{example}
  The computation of the SHAP scores for the explanations problems 
  in~\cref{fig:ex01,fig:ex02} are shown in~\cref{fig:svs01,fig:svs02},
  respectively.
\end{example}

\begin{figure*}[t]
  \centering
  \renewcommand{\tabcolsep}{0.725em}
  \renewcommand{\arraystretch}{0.975}
  \scalebox{0.975}{\begin{tabular}{cccccc} \toprule
  \multicolumn{6}{c}{$i=1$} \\
  \toprule
  $\fml{S}$ &
  $\cfn{e}(\fml{S})$ &
  $\cfn{e}(\fml{S}\cup\{1\})$ &
  $\Delta_1(\fml{S})$ &
  $\varsigma(\fml{S})$ &
  $\varsigma(\fml{S})\times\Delta_1(\fml{S})$
  \\
  \midrule[0.875pt]
  $\emptyset$ &
  $\nfrac{17}{12}$ &
  $1$ &
  $-\nfrac{5}{12}$ &
  $\sfrac{1}{3}$ &
  $-\nfrac{5}{36}$
  \\
  $\{2\}$ &
  $\nfrac{10}{12}$ &
  $1$ &
  $-\nfrac{2}{3}$ &
  $\sfrac{1}{6}$ &
  $-\nfrac{2}{18}$
  \\
  $\{3\}$ &
  $\nfrac{1}{2}$ &
  $1$ &
  $\nfrac{1}{2}$ &
  $\sfrac{1}{6}$ &
  $\nfrac{1}{12}$
  \\
  $\{2,3\}$ &
  $\nfrac{1}{2}$ &
  $1$ &
  $\nfrac{1}{2}$ &
  $\sfrac{1}{3}$ &
  $\nfrac{1}{6}$
  \\
  \midrule[0.75pt]
  \multicolumn{5}{r}{$\svn{e}(1)~~=$} & \multicolumn{1}{c}{0} \\
  \midrule[0.875pt]
  \multicolumn{6}{c}{$i=2$} \\
  \toprule
  $\fml{S}$ &
  $\cfn{e}(\fml{S})$ &
  $\cfn{e}(\fml{S}\cup\{2\})$ &
  $\Delta_2(\fml{S})$ &
  $\varsigma(\fml{S})$ &
  $\varsigma(\fml{S})\times\Delta_2(\fml{S})$
  \\
  \midrule[0.875pt]
  $\emptyset$ &
  $\nfrac{17}{12}$ &
  $\nfrac{5}{3}$ &
  $\nfrac{1}{4}$ &
  $\sfrac{1}{3}$ &
  $\nfrac{1}{12}$
  \\
  $\{1\}$ &
  $1$ &
  $1$ &
  $0$ &
  $\sfrac{1}{6}$ &
  $0$
  \\
  $\{3\}$ &
  $\nfrac{1}{2}$ &
  $\nfrac{1}{2}$ &
  $0$ &
  $\sfrac{1}{6}$ &
  $0$
  \\
  $\{1,3\}$ &
  $1$ &
  $1$ &
  $0$ &
  $\sfrac{1}{3}$ &
  $0$
  \\
  \midrule[0.75pt]
  \multicolumn{5}{r}{$\svn{e}(2)~~=$} & \multicolumn{1}{c}{0.08(3)} \\
  \midrule[0.875pt]
  \multicolumn{6}{c}{$i=3$} \\
  \toprule
  $\fml{S}$ &
  $\cfn{e}(\fml{S})$ &
  $\cfn{e}(\fml{S}\cup\{3\})$ &
  $\Delta_3(\fml{S})$ &
  $\varsigma(\fml{S})$ &
  $\varsigma(\fml{S})\times\Delta_3(\fml{S})$
  \\
  \midrule[0.875pt]
  $\emptyset$ &
  $\nfrac{17}{12}$ &
  $\nfrac{1}{2}$ &
  $-\nfrac{11}{12}$ &
  $\sfrac{1}{3}$ &
  $-\nfrac{11}{36}$
  \\
  $\{1\}$ &
  $1$ &
  $1$ &
  $0$ &
  $\sfrac{1}{6}$ &
  $0$
  \\
  $\{2\}$ &
  $\nfrac{5}{3}$ &
  $\nfrac{1}{2}$ &
  $-\nfrac{7}{6}$ &
  $\sfrac{1}{6}$ &
  $-\nfrac{7}{36}$
  \\
  $\{1,2\}$ &
  $1$ &
  $1$ &
  $0$ &
  $\sfrac{1}{3}$ &
  $0$
  \\
  \midrule[0.75pt]
  \multicolumn{5}{r}{$\svn{e}(3)~~=$} & \multicolumn{1}{c}{$-0.5$} \\
  \bottomrule[1pt]
\end{tabular}
}
  \caption{Computation of SHAP scores for $\fml{E}_{1}$.}
  \label{fig:svs01}
\end{figure*}

\begin{figure*}[t]
  \smallskip
  \centering
  \renewcommand{\tabcolsep}{0.725em}
  \renewcommand{\arraystretch}{0.975}
  \scalebox{0.975}{\begin{tabular}{cccccc} \toprule
  \multicolumn{6}{c}{$i=1$} \\
  \toprule
  $\fml{S}$ & $\cf_e(\fml{S})$ & $\cf_e(\fml{S}\cup\{1\})$ &
  $\Delta_1(\fml{S})$ & $\varsigma(\fml{S})$ &
  $\varsigma(\fml{S})\times\Delta_1(\fml{S})$ \\
  \midrule[0.875pt]
  $\emptyset$ & $\nfrac{3}{4}$ & 1 & $\nfrac{1}{4}$ & $\nfrac{1}{2}$ &
  $\nfrac{1}{8}$ \\
  $\{2\}$ & $\nfrac{5}{4}$ & 1 & $\nfrac{-1}{4}$ & $\nfrac{1}{2}$ &
  $\nfrac{-1}{8}$ \\
  \midrule
  \multicolumn{5}{r}{$\svn{e}(1)~~=$} & \multicolumn{1}{c}{0} \\
  \midrule[0.875pt]
  \multicolumn{6}{c}{$i=2$} \\
  \toprule
  $\fml{S}$ & $\cfn{e}(\fml{S})$ & $\cfn{e}(\fml{S}\cup\{2\})$ &
  $\Delta_2(\fml{S})$ & $\varsigma(\fml{S})$ &
  $\varsigma(\fml{S})\times\Delta_2(\fml{S})$ \\
  \midrule[0.875pt]
  $\emptyset$ & $\nfrac{3}{4}$ & $\nfrac{5}{4}$ & $\nfrac{1}{2}$ & $\nfrac{1}{2}$
  & $\nfrac{1}{4}$ \\
  $\{1\}$ & 1 & 1 & 0 & $\nfrac{1}{2}$ & 0 \\
  \midrule[0.75pt]
  \multicolumn{5}{r}{$\svn{e}(2)~~=$} & \multicolumn{1}{c}{$0.25$} \\
  \bottomrule
\end{tabular}
}
  \caption{Computation of SHAP scores for $\fml{E}_{2}$.}
  \label{fig:svs02}
\end{figure*}

\paragraph{Computing Shapley values \& SHAP scores.}
%
Motivated by the complexity of their exact computation, Shapley values
are usually approximated. This is the case with the tool SHAP, but
there exist well-known alternatives~\cite{tejada-cor09}.
For the case of SHAP scores, exact algorithms have been
studied~\cite{barcelo-jmlr23}.
Finally, for the special case of weighted voting games, there exist
pseudo-polynomial time algorithms based on dynamic
programming, e.g.~\cite{matsui-jorsj00}. 

\paragraph{The similarity predicate.}
%
With the purpose of enabling a unified treatment of logic-based XAI,
we can abstract away some details of ML models~\cite{ms-isola24}.
Given an ML model and some input $\mbf{x}$, the computed prediction is
\emph{distinguishable} with respect to the sample $(\mbf{v},q)$ if the
observed change in the model's output is deemed sufficient; 
otherwise it is \emph{similar} (or indistinguishable).
This is represented by a \emph{similarity} predicate (which can be
viewed as a boolean function) 
$\similar:\mbb{F}\to\{\bot,\top\}$ (where $\bot$ signifies
\emph{false}, and $\top$ signifies \emph{true}). Concretely,
$\similar(\mbf{x};\fml{E})$ holds true iff the change in the ML model
output is deemed \emph{insufficient} and so no observable difference
exists between the ML model's output for $\mbf{x}$ and $\mbf{v}$.%
\footnote{
Throughout the paper, parameterization are shown after the separator
';', and will be elided when clear from the context.}
For regression problems, we write instead $\similar$ as the
instantiation of a template predicate,
i.e.\ $\similar(\mbf{x};\fml{E})=\tsimilar(\mbf{x};\fml{E},\delta)$,
where $\delta$ is an optional measure of output change, which can be
set to 0.%
\footnote{%
Exploiting a threshold to decide whether there exists an observable
change has been used in the context of adversarial
robustness~\cite{barrett-nips23}. Furthermore, the relationship
between adversarial examples and explanations is
well-known~\cite{inms-nips19,barrett-nips23,msh-cacm24}.}

For regression problems, we represent relevant changes to the output
by a parameter $\delta$. Given a change in the input from $\mbf{v}$ to
$\mbf{x}$, a change in the output is indistinguishable (i.e.\ the
outputs are similar) if,
\begin{equation} \label{eq:deftsim}
  \similar(\mbf{x};\fml{E}) 
  \coloneq \tsimilar(\mbf{x};\fml{E},\delta)
  \coloneq [|\pi(\mbf{x})-\pi(\mbf{v})|\le\delta]
\end{equation}
otherwise, it is distinguishable.

For classification problems, similarity is defined to equate with not
changing the predicted class. Given a change in the input from 
$\mbf{v}$ to $\mbf{x}$, a change in the output is indistinguishable
(i.e.\ the outputs are similar) if,
\begin{equation} \label{eq:defsim}
  \similar(\mbf{x};\fml{E})\coloneq[\pi(\mbf{x})=\pi(\mbf{v})]
\end{equation}
otherwise, it is distinguishable.

\begin{example}
  For the two running examples (see~\cref{fig:ex01,fig:ex02}), it
  suffices to use~\cref{eq:defsim} as the similarity predicate,
  i.e.\ it suffices to test equality of prediction.
\end{example}

\section{Logic-Based Explainability} \label{sec:fxai}

In this paper, we study both logic-based explanations obtained from
some logic representation of the ML model, but also from some sample
of the ML model's behavior. The former are referred to as model-aware
explanations, whereas the latter are referred to as model-agnostic
explanations. Model-agnostic explanations represent a rigorous
alternative to the explanations obtained with tools such as
Anchors~\cite{guestrin-aaai18}.
Throughout, it is assumed an explanation problem $\fml{E}=(M,I)$,
where $M=(\fml{F},\mbb{F},\mbb{V},\pi)$ is an ML model, and
$I=(\mbf{v},p)$ is an instance.

\subsection{Model-Aware Explanations}

Two types of logic-based explanations have been studied:
abductive~\cite{inms-aaai19} and contrastive~\cite{inams-aiia20}.
The following paragraphs provide brief overviews of their definitions.

\paragraph{Abductive explanations.}
%
Given an instance $(\mbf{v},p)$, a weak abductive explanation (WAXp)
is a set $\fml{X}\subseteq\fml{F}$ such that,
\begin{equation} \label{eq:waxp}
  \waxp(\fml{X};\fml{E}) \quad \coloneq \quad
  \forall(\mbf{x}\in\mbb{F}).\left[\mbf{x}_{\fml{X}}=\mbf{v}_{\fml{X}}\right]\limply\left[\similar(\mbf{x};\fml{E})\right]
\end{equation}
Thus, if the features in $\fml{X}$ are fixed to the values dictated by
$\mbf{v}$, then the prediction is guaranteed to be $\pi(\mbf{v})=p$.
A WAXp $\fml{X}\subseteq\fml{F}$ is an \emph{abductive explanation}
(AXp) if it is subset-minimal, i.e.\ none of its subsets is a WAXp.
A predicate $\axp$ is associated with the condition of a set of
features being an AXp.
Although the relationship between the definition of AXp and
logic-based abduction~\cite{gottlob-jacm95} might seem
straightforward, it has been the subject of emergent
controversy~\cite{sharygina-cav25}.
\cref{app:abduct} provides a simple argument detailing the connection
between AXps and logic-based abduction.

\begin{example}
  For $\fml{E}_1=(M_1,I_1)$ (see~\cref{fig:ex01}), inspection of the
  DT confirms that $\fml{X}_1=\{1\}$ is a WAXp, i.e.\ fixing the value
  of feature 1 suffices for guaranteeing that the prediction is 1.
  In this case, it is also plain that $\fml{X}_1$ is an AXp, since
  $\emptyset$ cannot be an AXp (as we assume non-constant ML models).
  \\
  For $\fml{E}_2=(M_2,I_2)$ (see~\cref{fig:ex02}), inspection of the
  RT confirms that $\fml{X}_2=\{1\}$ is a WAXp, for the same reasons
  as above. Also, it is the case that $\fml{X}_2$ is an AXp.
\end{example}

\paragraph{Contrastive explanations.}
%
Given an instance $(\mbf{v},p)$, a weak contrastive explanation (WCXp)
is a set $\fml{Y}\subseteq\fml{F}$ such that,
\begin{equation} \label{eq:wcxp}
  \wcxp(\fml{Y};\fml{E}) \quad \coloneq \quad
  \exists(\mbf{x}\in\mbb{F}).\left[\mbf{x}_{\fml{F}\setminus\fml{Y}}=\mbf{v}_{\fml{F}\setminus\fml{Y}}\right]\land\left[\neg\similar(\mbf{x};\fml{E})\right]
\end{equation}
Thus, when the features in $\fml{Y}$ are allowed to take some value from
their domains (and the remaining features are fixed to the values
dictated by $\mbf{v}$), it is the case that the prediction can be made
different from $\pi(\mbf{v})=p$.
A WCXp $\fml{Y}\subseteq\fml{F}$ is a \emph{contrastive explanation}
(CXp) if it is subset-minimal, i.e.\ none of its subsets is a WCXp.
A predicate $\cxp$ is associated with the condition of a set of
features being a CXp. 

Contrastive explanations are tightly related with adversarial
examples~\cite{goodfellow-iclr15} (AExs), namely when the measure of
distance is $l_0$, i.e.\ the Hamming distance, and
the target are subset-minimal and not cardinality-minimal sets of
features~\cite{inms-nips19,barrett-nips23,msh-cacm24}.
Thus, any (W)CXp represents an AEx, since it denotes an example of the
features to change so that the prediction also changes.

Moreover, it is well-known that AXps are minimal hitting sets (MHSes)
of the CXps and vice-versa~\cite{inams-aiia20}. MHS duality builds on
Reiter's seminal work on model-based diagnosis in the
1980s~\cite{reiter-aij87}, and plays a key role in the enumeration of
explanations~\cite{inams-aiia20,ms-rw22}.

\begin{example}
  For $\fml{E}_1=(M_1,I_1)$ (see~\cref{fig:ex01}), inspection of the
  DT confirms that $\fml{Y}_1=\{1\}$ is a WCXp, i.e.\ allowing the
  value of feature 1 to change allows changing the prediction to a
  value other than 1.
  In this case, it is also plain that $\fml{Y}_1$ is a CXp, since
  $\emptyset$ cannot be an CXp (i.e.\ if all features are fixed, the
  prediction cannot change).
  \\
  For $\fml{E}_2=(M_2,I_2)$ (see~\cref{fig:ex02}), inspection of the
  RT confirms that $\fml{Y}_2=\{1\}$ is a WCXp, for the same reasons
  as above. Also, it is the case that $\fml{Y}_2$ is a CXp.
  \\
  Furthermore, one can compute all the CXps for the running examples
  using the polynomial-time algorithm proposed in earlier
  work~\cite{hiims-kr21}. In contrast, the set of AXps is obtained by
  hitting set
  dualization~\cite{khachiyan-jalg96,liffiton-jar08}. (Observe that,
  since all the CXps are computed in polynomial time, there is no need
  to exploit implicit hitting set dualization~\cite{lpmms-jar16}.)
  For the two running examples, the set of CXps is $\{\{1\}\}$ and the
  set of AXps is $\{\{1\}\}$.
  For both $\fml{E}_1$ and $\fml{E}_2$, the sole adversarial example
  requires changing the value of feature 1; there is no other way to
  change the prediction of 1 other than changing the value of feature
  1.
\end{example}

Finally, a feature is said to be \emph{relevant} if it occurs in some
AXp. Otherwise, the feature is \emph{irrelevant}. A relevant feature
also occurs in some CXp, whereas an irrelevant feature does
not~\cite[Proposition~8]{ms-rw22}.
If one computes all the AXps (or all the CXps), then feature relevancy
is decided in linear time on the set of AXps (or CXps).

\begin{example}
  For $\fml{E}_1=(M_1,I_1)$ (see~\cref{fig:ex01}), it is simple to
  conclude that feature 1 is relevant and that the features 2 and 3
  are irrelevant~\cite{msh-cacm24}.
\end{example}

\paragraph{Progress, status \& assessment.}
%
Logic-based explainability has been the subject of significant
progress since the initial works in
2018/19~\cite{darwiche-ijcai18,inms-aaai19}.
Recent overviews provide a detailed account of this
progress~\cite{ms-rw22,darwiche-lics23,ms-isola24}.
For several families of classifiers, it has been shown that there
exist polynomial-time algorithms for computing a single
AXp/CXp~\cite{msgcin-nips20,msgcin-icml21,hiims-kr21,cms-cp21,hiicams-aaai22,cms-aij23,ccms-kr23,barcelo-pods25}.
For other families of classifiers, it has been shown that finding
one AXp/CXp is computationally hard, and different logic encodings
have been
devised~\cite{inms-aaai19,ims-sat21,ims-ijcai21,iisms-aaai22}.
\cref{app:abduct} outlines the method that is used in practice when
the computation of AXps/CXps involves logic encodings.
Alternative methods based on compilation to a canonical representation
have also been studied in the
past~\cite{darwiche-lics23,steffen-tmf25}.

\subsection{Model-Agnostic Explanations}


In contrast to the model-aware case, model-agnostic explanations are
defined with respect to a given sample of the ML model's behavior.
Different methods can be envisioned to crease such a sample, including
those used by tools such as LIME~\cite{guestrin-kdd16},
SHAP~\cite{lundberg-nips17} and Anchors~\cite{guestrin-aaai18}, but
also the dataset used for training the ML model.
A sample $\mbb{S}$ is a tuple $(\mbb{D},\mbf{p}^T)$, where $\mbb{D}$
is a $n\times{m}$ matrix, such that each row $j$ denotes a point
$\mbf{d}_j$ of $\mbb{F}$. In addition, $\mbf{p}=(p_1,\ldots,p_n)$,
such that $p_j\in\mbb{V},j=1,\ldots,n$, and such that
$\pi(\mbf{d}_j)=p_j,j=1,\ldots,n$.

The definition of model-agnostic explanations mimic those of the
model-aware case, but now quantification is restricted to the points
in the sample. Thus,
\begin{equation} \label{eq:waxp2}
  \waxp(\fml{X};\fml{E}) \quad \coloneq \quad
  \forall(\mbf{x}\in\mbb{D}).\left[\mbf{x}_{\fml{X}}=\mbf{v}_{\fml{X}}\right]\limply\left[\similar(\mbf{x};\fml{E})\right]
\end{equation}
and,
\begin{equation} \label{eq:wcxp2}
  \wcxp(\fml{Y};\fml{E}) \quad \coloneq \quad
  \exists(\mbf{x}\in\mbb{D}).\left[\mbf{x}_{\fml{F}\setminus\fml{Y}}=\mbf{v}_{\fml{F}\setminus\fml{Y}}\right]\land\left[\neg\similar(\mbf{x};\fml{E})\right]
\end{equation}
(W)AXps/(W)CXps for model-aware and model-agnostic are referred to
using the same predicate names; the difference will be clear from the 
context.
Recent works have studied rigorous model-agnostic
explanations~\cite{cooper-ecai23,amgoud-ecai24,msllm-ijcai25}.

\section{The Flaws of SHAP Scores} \label{sec:svmyth}

Several flaws in the theory underlying SHAP scores have been reported
since 2023~\cite{hms-corr23a,hms-corr23b,hms-corr23c,msh-corr23}.
Since then, further results confirmed the flaws of theoretical SHAP
scores~\cite{msh-cacm24,hms-ijar24,lhms-corr24b}.
Concretely, it has been shown that there arbitrarily many ML models
for which the theoretical scores will mislead human decision makers.
This section briefly overviews the known flaws of theoretical SHAP
scores.
Throughout this section, we focus on cases where the computed SHAP
score is clearly \emph{misleading}, i.e.\ importance is given to
features that are manifestly unimportant, and no importance is given
to features that are guaranteed to be important.

\begin{table}[t]
  \caption{Theoretical SHAP scores vs.~feature relevancy for the
    running examples (see~\cref{fig:ex01,fig:ex02}).}
  \label{tab:cmp}
  \smallskip
  \centering
  \renewcommand{\tabcolsep}{0.75em}
  \renewcommand{\arraystretch}{1.15}
  \begin{tabular}{ccc|ccc} \toprule
    \multicolumn{3}{c}{$\fml{E}_1=(M_1,I_1)$} &
    \multicolumn{3}{c}{$\fml{E}_2=(M_2,I_2)$} \\
    \toprule
    Feature $i$ & $\svn{e}(i)$ & \multicolumn{1}{c}{Relevant?} &
    Feature $i$ & $\svn{e}(i)$ & \multicolumn{1}{c}{Relevant?} \\
    \toprule
    1 & 0 & Yes & 1 & 0 & Yes \\
    2 & 0.08(3) & No & 2 & $0.25$ & No \\
    3 & $-0.5$ & No & --- & --- & --- \\
    \bottomrule
  \end{tabular}
\end{table}

\paragraph{Classification \& regression models.}
%
For the two running examples (see~\cref{fig:ex01,fig:ex02}), feature 1
is relevant and the remaining features are irrelevant. However, as
shown in~\cref{fig:svs01,fig:svs02}, the SHAP score for the relevant
feature 1 is 0, which
signifies~\cite{kononenko-jmlr10,kononenko-kis14} \emph{no importance
for the prediction}. In contrast, the SHAP score for the remaining
features is not 0, which signifies \emph{some importance for the
prediction}. Therefore, for the two running examples, the relative
importance of features obtained using SHAP scores is evidently
misleading for a human decision maker. \cref{tab:cmp} summarizes the
comparison between computed theoretical SHAP scores and feature
relevancy.

\paragraph{Continuous \& differentiable models.}
%
A possible criticism regarding classification models and regression
models with discrete domains and discrete predicted values is that
these may not represent all the possible ranges of uses of ML.
However, one can create examples where both the features and the
predicted values take (uncountable) real values.
The following example, first discussed in~\cite{lhms-corr24b},
confirms that the flaws of SHAP scores also exist in the case of
real-valued ML models. In this case, the definition of the similarity
predicate is from~\eqref{eq:deftsim}, and so a value of $\delta$ must
be chosen. We will pick some $\delta<\sfrac{1}{4}$.

\begin{example}(Regression model $M_3$.) \label{ex:rm02}
  We consider a regression problem defined over two real-valued
  features, taking values from interval
  $[-\sfrac{1}{2},\sfrac{3}{2}]$.
  Thus, we have $\fml{F}_3=\{1,2\}$,
  $\mbb{D}_{3,1}=\mbb{D}_{3,2}=\mbb{D}_3=[-\sfrac{1}{2},\sfrac{3}{2}]$,
  $\mbb{F}_3=\mbb{D}_3\times\mbb{D}_3$.
  (We also let $\mbb{D}^{+}_{3}=[\sfrac{1}{2},\sfrac{3}{2}]$
  and $\mbb{D}^{-}_{3}=\mbb{D}_3\setminus\mbb{D}^{+}_{3}$.)
  %
  In addition, the regression model maps to real values,
  i.e.\ $\mbb{V}_3=\mbb{R}$, and is defined as follows:
  \[
    \pi_3(x_1,x_2) =
    \left\{
    \begin{array}{lcl}
      x_1 & ~~ & \tn{if $x_1\in\mbb{D}^{+}_{3}$} \\ 
      x_2-2 & ~~ & \tn{if
        $x_1\not\in\mbb{D}^{+}_{3}\land{x_2}\not\in\mbb{D}^{+}_{3}$}\\
      x_2+1 & ~~ & \tn{if
        $x_1\not\in\mbb{D}^{+}_{3}\land{x_2}\in\mbb{D}^{+}_{3}$}\\
    \end{array}
    \right.
  \]
  As a result, the regression model is represented by
  $M_3=(\fml{F}_3,\mbb{F}_,\mbb{V}_3,\pi_3)$.
  Moreover, we assume the target instance to be
  $I_3=(\mbf{v}_3,p_3)=((1,1),1)$,
  and so the explanation problem becomes
  $\fml{E}_3=(M_3,(\mbf{v}_3,p_3))$.
\end{example}

\begin{table}[t]
  \caption{Expected values of $\pi_3$, for each possible set
    $\fml{S}$ of fixed features, and given the sample $((1,1),1)$.
  These expected values also apply in the case of $\pi_4$
  (see~\cref{fig:rmlc}).}
  \label{tab:avg03}
  \smallskip
  \centering
  \renewcommand{\tabcolsep}{0.775em}
  \begin{tabular}{ccccc} \toprule
    $\fml{S}$ & $\emptyset$ & $\{1\}$ & $\{2\}$ & $\{1,2\}$ \\
    \midrule[0.75pt]
    $\exv[\pi_3(\mbf{x})\,|\,\mbf{x}_{\fml{S}}=\mbf{v}_{\fml{S}}]$
    & $\sfrac{1}{2}$ & 1 & $\sfrac{3}{2}$ & 1 \\
    \bottomrule
  \end{tabular}
\end{table}

\begin{example}(AXps, CXps and AExs for $\fml{E}_3$.)
  Given the regression model $M_3$, we define the similarity
  predicate by picking a suitably small value $\delta$, e.g.\ 
  $\delta<\sfrac{1}{4}$ as suggested above.
  This suffices to ensure that the similarity predicate $\similar$
  only takes value $\top$ when feature 1 takes value 1.\\
  Given the above, the similarity predicate takes value $\top$ only
  when feature 1 takes value 1, and independently of the value
  assigned to feature 2.\\
  Thus, fixing feature 1 ensures that the similarity predicate always
  takes value $\top$. Otherwise, if feature 1 is allowed to take a
  value other than 1, then the similarity predicate can take value
  $\bot$, and so the WAXp condition does not hold.
  As a result, $\{1\}$ is one (and the only) AXp, and $\{1\}$ is also
  one (and the only) CXp.\\
  A similar analysis allow concluding that a $l_0$-minimal AEx
  exists iff feature 1 is allowed to take any value from
  its domain.
\end{example}

\begin{example}(SHAP scores for $\fml{E}_3$.)
  The expected values of $\pi_3$ for all possible sets $\fml{S}$ of
  fixed features is shown in~\cref{tab:avg03}.%
  \footnote{%
  The computation of these expected values is fairly straightforward,
  and is summarized in the supplemental materials.}
  %
  Hence, the computation of SHAP scores is also the one shown
  in~\cref{fig:svs02}, given the expected values.
  As a result, $\svn{e}(1)=0$ and $\svn{e}(2)=\sfrac{1}{2}$.
\end{example}

\begin{figure*}[t]
  \centering
  \begin{tabular}{l}
    $  \pi_4(x_1,x_2) =
      \left\{
      \begin{array}{lcl}
        x_1 &
        \quad & \tn{if $x_2\le1\land\alpha{x_1}\le\alpha$} \\[2pt]
        (1+4|\alpha|)x_1-4|\alpha| &
        \quad & \tn{if $x_2\le1\land\alpha{x_1}\ge\alpha$} \\[2pt]
        28|\alpha|{x_1}{x_2} + (1-28|\alpha|)x_1 -28|\alpha|{x_2}+28|\alpha| &
        \quad & \tn{if $x_2\ge1\land\alpha{x_1}\le\alpha$} \\[2pt]
        -4|\alpha|{x_1}{x_2}+(1+8|\alpha|)x_1+4|\alpha|{x_2}-8|\alpha| &
        \quad & \tn{if $x_2\ge1\land\alpha{x_1}\ge\alpha$} \\[5pt]
      \end{array}
      \right.
      $
  \end{tabular}
  \caption{Example of regression model that is Lipschitz continuous.}
  \label{fig:rmlc}
\end{figure*}

Given the example above, we have constructed a regression example over
uncountable domains and predicted values, such that the SHAP scores
mislead.
Nevertheless, one possible criticism of the example above is that the
ML model is \emph{not} continuous, and one may argue that often used
ML models are continuous.
We now present a \emph{family} of continuous ML models, that
also produce misleading  SHAP scores. The family of models is
parameterized on $\alpha$. We will then argue that the proposed family
of ML models also ensures Lipschitz continuity.

\begin{example} \label{ex:rm04}
  We consider a regression problem defined over two real-valued
  features, taking values from interval $[0,2]$.
  Thus, we have $\fml{F}=\{1,2\}$,
  $\mbb{D}_{4,1}=\mbb{D}_{4,2}=\mbb{D}_4=[0,2]$,
  $\mbb{F}_4=\mbb{D}_4\times\mbb{D}_4$.
  In addition, the regression model maps to real values,
  i.e.\ $\mbb{V}_4=\mbb{R}$, and is defined as shown
  in~\cref{fig:rmlc}.
  The value of $\alpha$ is such that $\alpha\in\mbb{R}\setminus\{0\}$.
  We will pick $\alpha=\sfrac{1}{4}$.
  As a result, the regression model is represented by
  $M_4=(\fml{F}_4,\mbb{F}_4,\mbb{V}_4,\pi_4)$.
  Moreover, we assume the target sample to be
  $I_4=(\mbf{v}_4,p_4)=((1,1),1)$, and so the explanation problem
  becomes $\fml{E}_4=(M_4,I_4)$.
\end{example}

By inspection, it is plain that $\pi_4$ is continuous; this is
further discussed below.

\begin{example}(AXps, CXps and AExs for $\fml{E}_4$.)
  As before, it is plain to reach the conclusion that the set of AXps
  is $\{\{1\}\}$, and this is also the set of CXps. Moreover, and as
  before, there is a $l_0$-minimal AEx containing feature 1.
\end{example}

\begin{example}(SHAP scores for $\fml{E}_4$.)
  The regression model $M_4$ is devised such that the expected
  values of $\pi_4$ for each possible set $\fml{S}$ of 
  fixed features are exactly the ones shown in~\cref{tab:avg03}.
  As a result, the computed SHAP scores are the same as before, and so
  they are again misleading.
\end{example}

As shown in~\cite{lhms-corr24b}, $\pi_4$ is Lipschitz-continuous.
Thus, the continuity of $\pi_4$, which is claimed above, is implied
by the fact that $\pi_4$ Lipschitz-continuous.
Finally, recent work also argues that the flaws of SHAP scores can
also exist in arbitrarily differentiable ML
models~\cite{lhms-corr24b}.

\paragraph{Significance of SHAP's flaws.}
%
As summarized above, earlier work provided extensive evidence regarding 
the flaws of SHAP
scores~\cite{hms-corr23a,hms-corr23b,hms-corr23c,msh-corr23,msh-cacm24,hms-ijar24,lhms-corr24b},
that cover classification, regression, but also
(Lipschitz-)continuous and differentiable ML models.
The majority of cases relate with assigning importance to irrelevant
features, not assigning importance to relevant features, and also
allowing the relative importance of relevant and irrelevant features
to be reversed~\cite{msh-cacm24}. In turn, this can serve to mislead
human decision makers.
The fact that the tool SHAP approximates Shapley values only makes
matters worse. This is further discussed in~\cref{sec:res}.

Despite the mounting evidence regarding the flaws of SHAP scores, the
growing popularity of the tool SHAP~\cite{lundberg-nips17} is evident
and underlined by its impact.
For example, and quoting from recent work~\cite{he-phd25,he-ssrn25}:\\
%
\emph{``... 
we note that certain studies have pointed out the limitations of SHAP
(...).
However, it has been used extensively in the literature due to its
demonstrated effectiveness and has contributed to significant
discoveries in science, engineering, and business, which were
discussed in publications in leading journals such as Science, Nature
Machine Intelligence, Nature Communications, and Management Science
(see ...%
), just to name a few (the paper by Lundberg and Lee (2017)
has been cited more than 34,000 times as of May 2025).''}
\\
Unfortunately, rigor cannot be established by popularity.
The flaws of SHAP scores have been demonstrated with a vast number of 
case studies (and supporting theory), included in this and in earlier
papers~\cite{hms-corr23a,hms-corr23b,hms-corr23c,msh-corr23,msh-cacm24,hms-ijar24,lhms-corr24b}.
Some researchers have recently suggested the assessment of results by
domain experts or additional interpretation to confirm the quality 
of SHAP's
results~\cite{maposa-sr25,dewey-nt25,liu-aem25,vonluxburg-icml25pp}. Clearly,
for complex ML models, this is error-prone.
Given the above, and in light of the identified flaws, we contend that
the most significant conclusions drawn in the many publications that
build on the tool SHAP~\cite{lundberg-nips17} ought to be reassessed.

\section{Corrected SHAP Scores -- Theory} \label{sec:fixsv}

Given the theoretical flaws of SHAP scores, a natural question is
whether such flaws can be fixed.
This section proposes a game for explainability which is guaranteed
not to exhibit the limitations of the SHAP scores used in the tool
SHAP.

\subsection{Properties of Characteristic Functions for XAI}
\label{ssec:props}

Recent work~\cite{lhms-corr24a,lhms-aaai25} proposed several
properties that characteristic functions for XAI games should respect
so that the identified flaws of SHAP scores are eliminated.

\paragraph{Strong value independence.}
Let $\fml{M}_1=(\fml{F},\mbb{F},\fml{T}_1,\tau_1)$ be an ML model,
with domain $\mbb{D}_i$ for each feature $i\in\fml{F}$. 
Moreover, let $\fml{M}_2=(\fml{F},\mbb{F},\fml{T}_2,\tau_2)$ be
another classifier, with the same domains.
%
In addition, let $\mu:\fml{K}_1\to\fml{K}_2$ be a 
mapping from $\fml{T}_1$ to $\fml{T}_2$, such that for
$q\in\fml{T}_1$, 
and such that,
$\forall(b\in\fml{T}_1).[(b\not=q)\limply(\mu(b)\not=\mu(q))]$
%
%
Finally, let the target samples be $(\mbf{v},p)$, for $\fml{M}_1$,
and $(\mbf{v},\mu(p))$ for $\fml{M}_2$, thus defining the explanation
problems $\fml{E}_1=(\fml{M}_1,(\mbf{v},p))$ and
$\fml{E}_2=(\fml{M}_2,(\mbf{v},\mu(p)))$.
A characteristic function $\cfn{t}$ is \emph{strongly
value-independent} if, given $\mu$,
$\forall(i\in\fml{F}).[\svn{t}(i;\fml{E}_1)=\svn{t}(i;\fml{E}_2)]$
%
Given the above, the following result holds%

\paragraph{Compliance with feature (ir)relevancy.}
Characteristic functions should respect feature (ir)re\-le\-van\-cy,
i.e.\ a feature is irrelevant iff its (corrected) SHAP score is 0.
Formally, a characteristic function $\cf_t$ is compliant with feature 
(ir)re\-le\-van\-cy if,
%
\begin{equation} \label{eq:compliance}
  \forall(i\in\fml{F}).\irrelevant(i)\leftrightarrow(\svn{t}(i)=0)
\end{equation}
In previous work~\cite{hms-corr23a,hms-ijar24,msh-cacm24}, SHAP scores
are said to be \emph{misleading} when compliance with feature
(ir)relevancy is not respected. In the remainder of the paper, we
assign the same meaning to the term \emph{misleading}.

\paragraph{Numerical neutrality.}
Existing definitions of SHAP scores are based on expected values and
so require $\mbb{V}$ to be ordinal.
However, classification problems often contemplate categorical
classes. A characteristic function respects numerical neutrality if it
can be used with both numerical and non-numerical $\mbb{V}$.


\subsection{A Logic-Based Game for XAI}

%
Instead of defining the XAI game using a characteristic function
defined in terms of the expected value of the ML model, as has been
done in the
past~\cite{kononenko-jmlr10,kononenko-kis14,lundberg-nips17}, we
have proposed instead the following characteristic
function~\cite{lhms-corr24a,lhms-aaai25}:
\begin{equation} \label{eq:newnu}
  \cfn{a}(S;\fml{E}) \quad \coloneq \quad
  \left\{
  \begin{array}{lcl}
    1 & \quad & \tn{if~$\,\waxp(S;\fml{E})$} \\[3pt]
    0 & \quad & \tn{otherwise}\\
  \end{array}
  \right.
\end{equation}
(Clearly, the game is still defined on the set of features, with the
difference to earlier definitions being the characteristic function.)
As noted in earlier work~\cite{lhms-aaai25}, the relationship with the
characteristic functions used in a priori voting power are apparent.
It is also important to underline the connection between the proposed
game for XAI and logic-based explainability by feature selection.
Finally, the proposed characteristic function ensures that the
resulting game is a \emph{simple game}~\cite{elkind-bk12}, i.e.\ the
characteristic function is monotonically increasing, and takes values
from the set $\{0,1\}$.

Earlier work proved that the logic-based characteristic function
proposed above respects the properties listed in~\cref{ssec:props}
that characteristic functions for XAI should respect.


\section{Corrected SHAP Scores -- Practice}
\label{sec:nushap}

This section details the organization of nuSHAP~\cite{mshl-corr25},
a recently proposed alternative to the tool
SHAP~\cite{lundberg-nips17}.
Whereas the tool SHAP approximates the well-known SHAP scores, which
correspond to a game where the characteristic function is given
by~\eqref{eq:defsve}, nuSHAP uses the logic-based characteristic
function shown in~\eqref{eq:newnu}.

\paragraph{Computing Shapley values.}
%
Although polynomial-time algorithms have been proposed for computing
SHAP scores for restricted families of ML
models~\cite{barcelo-jmlr23}, those results do not apply in the case
of corrected SHAP scores, because the characteristic function tests
whether the given set is a WAXp and does not compute expected values.
At present, no polynomial-time algorithms have been devised for
computing corrected SHAP scores.
The solution that has been proposed in recent work~\cite{mshl-corr25}
is to adopt the well-known approximation algorithm of Castro,
G\'{o}mez and Tejada (CGT)~\cite{tejada-cor09}, which provides strong 
theoretical guarantees.
Thus, nuSHAP implements the CGT algorithm on the XAI game
$G_{a}=(\fml{F},\cfn{a})$.
The rest of this section discusses how the predicate $\waxp$ is
decided in practice.

\paragraph{Model-aware corrected SHAP scores.}
%
Assuming that we are equipped with an explainer capable of deciding
whether a set of features is a WAXp, then we can simply run the CGT
algorithm, at each step computing the characteristic function by
checking the predicate $\waxp$.
For several families of classifiers, this can be attained in
polynomial-time~\cite{ms-rw22}.

\paragraph{Model-agnostic corrected SHAP scores.}
%
For complex ML models, checking the predicate $\waxp$ cannot be
achieved in polynomial-time. For these cases, nuSHAP proposes the
computation of rigorous model-agnostic explanations, where the sample
is either obtained from the ML model, or alternatively by adopting the
dataset used for training the ML model.
One advantage of rigorous model agnostic explanations is that the
predicate $\waxp$ can be checked in
polynomial-time~\cite{cooper-ecai23,msllm-ijcai25}.

\paragraph{Open research topics.}
%
As noted above, the exact computation of corrected SHAP scores is
currently an open topic of research.
While the CGT algorithm~\cite{tejada-cor09} provides strong
theoretical guarantees, it is open whether alternatives might exist,
for either computing the SHAP scores approximately or exactly.

\section{Experiments} \label{sec:res}

%
This section summarizes the experiments comparing the tools
SHAP~\cite{lundberg-nips17} and nuSHAP~\cite{mshl-corr25}, a recently
proposed alternative.
The tool nuSHAP approximates the $\svn{a}$ SHAP scores, using the CGT
algorithm~\cite{tejada-cor09}. In contrast, the tool SHAP approximates
the $\svn{e}$ SHAP scores using its own approximation algorithm.
The goal of the experiments is to assess the quality of the SHAP
tool~\cite{lundberg-nips17} at ranking features in terms of their
relative importance for a given prediction.
To ensure a fair comparison, the sampling performed by the tool SHAP
was recorded. This sampling was then used by nuSHAP.

The tools SHAP\footnote{Available from~\url{https://github.com/slundberg/shap}.}
and nuSHAP were assessed on several well-known
classifiers~\cite{zhou-bk21}, namely: 
logistic regression (LR), 
decision tree (DT), 
$k$-nearest neighbors ($k$NN) classifier, 
boosted trees (BT), 
and Convolutional Neural Network (CNN).
LR, DT, and $k$NN models are trained using
scikit-learn~\cite{pedregosa2011scikit},
BT models are trained using the XGBoost algorithm~\cite{chen2016xgboost},
while CNN models are trained using TensorFlow\footnote{\url{https://www.tensorflow.org/}.}.
The comparison was conducted across a range of widely used tabular 
classification datasets selected from the PMLB
benchmark~\cite{Olson2017PMLB},
as well as the MNIST dataset~\cite{deng2012mnist} of handwritten
digits (0–9).
The classifiers and datasets used in the experiments are briefly
summarized in~\cref{tab:rbo_sum}. (Additional detail is included
in~\cite{mshl-corr25}.)
%
%
For each tabular dataset, we randomly picked 50 tested instances for
computing nuSHAP and SHAP scores.
For the MNIST dataset, we randomly selected 20 test instances to
compute these scores.
Moreover, we chose different model-agnostic SHAP explainers when
computing SHAP scores.
\emph{ExactExplainer} was used for the first and second sets,
\emph{PermutationExplainer} was used for the third set,
and \emph{SamplingExplainer} was used for the fourth and last set.
All explainers are provided with the entire training data so that the
SHAP tool can draw samples from it.
However, for MNIST a reduced training data was used, to curb the size
of sampling.
For the nuSHAP tool, the parameters of the CGT algorithm used were
$\epsilon=0.0015$ and $\alpha=0.015$, for all the tested instances.

\begin{table*}[t] 
  \caption{Summary of RBO values for all the tested instances, including the minimum, maximum, and mean RBO values.}
  \label{tab:rbo_sum}
  \smallskip
  \centering
  \renewcommand{\tabcolsep}{0.2125em}
  \renewcommand{\arraystretch}{1.075}
  \scalebox{0.875}{
    \begin{tabular}{llrrrrrrrrrrr}
      \toprule
      \multicolumn{2}{c}{\quad} &
      \multicolumn{2}{c}{LR} &
      \multicolumn{2}{c}{DT} &
      \multicolumn{3}{c}{$k$NN} &
      \multicolumn{3}{c}{BT} &
      \multicolumn{1}{c}{CNN}
      \\
      \cline{3-13}
      &           & adult & corral & iris & mux6 & conn..4 & spamb. & spectf & clean1 & coil..0 & dna  & MNIST \\
      \midrule
      \parbox[t]{2mm}{\multirow{2}{*}{\rotatebox[origin=c]{90}{Min}}}
      & nuSHAP$\;$vs$\;$SHAP      & 0.08  & 0.17   & 0.31 & 0.32 & 0.0       & 0.01     & 0.0    & 0.0    & 0.0      & 0.0  & 0.0 \\
      & nuSHAP$\;$vs$\;$|SHAP| & 0.05  & 0.12   & 0.27 & 0.32 & 0.0       & 0.05     & 0.0    & 0.0    & 0.0      & 0.03 & 0.0 \\
      \midrule
      \parbox[t]{2mm}{\multirow{2}{*}{\rotatebox[origin=c]{90}{Max}}}
      & nuSHAP$\;$vs$\;$SHAP      & 0.96  & 0.96   & 0.94 & 0.97 & 0.9       & 0.94     & 0.91   & 0.69   & 0.69     & 0.88 & 0.06 \\
      & nuSHAP vs.~|SHAP| & 0.88  & 0.97   & 0.94 & 0.95 & 0.77      & 0.94     & 0.91   & 0.88   & 0.69     & 0.88 & 0.06 \\
      \midrule
      \parbox[t]{2mm}{\multirow{2}{*}{\rotatebox[origin=c]{90}{Mean}}}
      & nuSHAP$\;$vs$\;$SHAP      & 0.37  & 0.53   & 0.84 & 0.7  & 0.21      & 0.41     & 0.2    & 0.12   & 0.05     & 0.17 & 0.0 \\
      & nuSHAP vs.~|SHAP| & 0.31  & 0.5    & 0.84 & 0.69 & 0.19      & 0.42     & 0.19   & 0.17   & 0.08     & 0.43 & 0.0 \\
      \bottomrule
    \end{tabular}
  }
\end{table*}

\begin{table*}[t] 
  \caption{Average runtime (in seconds) for computing SHAP and nuSHAP
    scores.
  }
  \label{tab:time_tabular}
  \smallskip
  \centering
  \scalebox{0.875}{
    \begin{tabular}{lrrrrrrrrrrr}
      \toprule
      & adult & corral & iris & mux6 & conn..4 & spamb. & spectf & clean1 & coil..0 & dna & MNIST \\
      \midrule
      SHAP & 3.4 & \textbf{0.1} & \textbf{0.0} & \textbf{0.0} & 21.7 & \textbf{0.5} & \textbf{0.7} & 6.8 & 28.2 & 23.0 & 281.3 \\
      nuSHAP & \textbf{1.9} & 1.5 & 1.5 & 1.5 & \textbf{4.5} & 2.7 & 2.9 & \textbf{2.7} & \textbf{1.7} & \textbf{4.5} & \textbf{48.9} \\
      \bottomrule
    \end{tabular}
  }
\end{table*}

SHAP and nuSHAP were compared using two metrics: i) the ranking of
feature importance imposed by different scores;
and ii) the runtime for computing these scores.
Specifically, for each tested instance, we first compute its nuSHAP
and SHAP scores.
We then determined the order of feature importance based on these
scores.
For SHAP, we considered two orders: one based on the original SHAP
scores and the other based on the absolute values of the SHAP scores. 
Next, we compare the order imposed by nuSHAP scores with the order
imposed by original SHAP scores, and separately compare the order
imposed by nuSHAP scores with the order imposed by the absolute values
of SHAP scores.

To compare the rankings of feature importance,
we used the metric rank-biased overlap
(RBO)~\cite{webber2010similarity} for each pair of scores.
RBO is a metric used to measure the similarity between two ranked
lists, and it ranges between 0 and 1.
A higher RBO value indicates a greater degree of similarity between
the two rankings, with 1 denoting a perfect match for the top-ranked
elements considered.
A publicly available
implementation\footnote{\url{https://github.com/changyaochen/rbo}.} of
RBO was used in our experiments.
Given the fact that human decision makers are able to grasp a reduced
number of features~\cite{miller-pr56}, we set \emph{persistence} to
$0.5$ and \emph{depth} to $5$ in our setting. This means that we
focused on the top-5 features while placing greater emphasis on the
top-3 features.
\cref{tab:rbo_sum} summarizes the RBO values for all the tested
instances, including the minimum, maximum, and mean RBO values.
As can be observed, there is essentially \emph{no} correlation between
the results of the tools nuSHAP and SHAP.
Finally, \cref{tab:time_tabular} presents the average running times
for computing the different scores, confirming that nuSHAP performs
similarly to SHAP.

\section{Conclusions} \label{sec:conc}

This paper provides a brief account of the ongoing efforts to develop
a rigorous alternative to the tool SHAP~\cite{lundberg-nips17}, which
is used ubiquitously as a method of feature attribution in XAI.
The paper revisits the limitations of the theory underlying SHAP, and
proposes a rigorous alternative, i.e.\ the so-called corrected SHAP
scores. Furthermore, the paper outlines possible methods for
computing corrected SHAP scores.
The experiments provide additional evidence regarding the limitations
of the tool SHAP.
Future work will target the exact computation of corrected SHAP
scores.

\paragraph{Acknowledgments.}
This work is supported by the Spanish Government grant
PID2023-152814OB-I00.

\newtoggle{mkbbl}

\settoggle{mkbbl}{false}

\bibliographystyle{plain}

\iftoggle{mkbbl}{
  \bibliography{refs}
}{

}

\appendix
\renewcommand{\theHsection}{A\arabic{section}}

\section{Abductive Explanations \& Logic-Based Abduction}
\label{app:abduct}

Throughout this section, we adopt the definition of (subset-minimal)
logic-based abduction from earlier work~\cite{gottlob-jacm95}.
Furthermore, and for simplicity, we assume an ML model with boolean
features and boolean prediction function, such that it has a
propositional logic representation.
Nevertheless, more expressive logics, e.g.\ (fragments of) first order 
logic, could be considered.
Let $T$ denote the logic encoding of some ML model $\mathcal{M}$. For
simplicity, we will assume that $T$ is a propositional formula,
represented in CNF, that the features are Boolean and that
$\mathbb{K}=\{\top,\bot\}$. The generalization to more expressive
logics is straightforward.
$T$ is defined on two sets of propositional variables
$X\cup{Y}=\{x_1,\ldots,x_m\}\cup\{y_1,\ldots,y_M\}$, where
$X=\{x_1,\ldots,x_m\}$ denotes the propositional variables associated
with the features of the ML model $\mathcal{M}$, and
$Y=\{y_1,\ldots,y_M\}$ denotes auxiliary variables. One variable,
e.g.\ $y_M$, is distinguished to denote the class computed by the
ML model.
For any assignment to the variables in
$\{x_1,\ldots,x_m\}$, representing some point $\mbf{x}\in\mbb{F}$,
there exists at least one assignment to the variables in
$\{y_1,\ldots,y_M\}$, such that $T$ is satisfiable, and such that
$y_M=\kappa(\mbf{x})$, and there are no assignments to the variables
in $\{y_1,\ldots,y_M\}$, such that $T$ is satisfiable and
$y_M\not=\kappa(\mbf{x})$.
Furthermore, we define $h_i\leftrightarrow(x_i\leftrightarrow{v_i})$
and
$o\leftrightarrow(\kappa(\mathbf{x})\leftrightarrow\kappa(\mathbf{v}))$. (Converting
these constraints to clausal form is straightforward. In this context,
$\kappa(\mbf{v})$ is a constant, and $\kappa(\mbf{x})$ denotes the
value assigned to the corresponding variable $y_M$ in $Y$.)
Now, we define
$R\triangleq\left[{T}\land_{i=1}^m\left({h_i\leftrightarrow(x_i\leftrightarrow{v_i})}\right)\land(o\leftrightarrow(\kappa(\mathbf{x})\leftrightarrow\kappa(\mathbf{v})))\right]$,
and let
$H=\{h_1,\ldots,h_m\}$ denote the set
of hypotheses, and let $M={o}$ denote the manifestation.
Furthermore, let $V=\vars(R)$ denote the set of propositional
variables used in the definition of $R$.
Finally, the tuple $P=\langle{V},H,M,R\rangle$ defines a
(propositional) logic-based abduction problem~\cite{gottlob-jacm95},
such that an explanation (or solution) for the abduction problem is a
set $X\subseteq{H}$ such that:
(i) $R\cup{X}$ is consistent; and (ii) $R\cup{X}\entails{M}$.
%
%
Clearly, for any set $X\subseteq{H}$, it is the case that $R\cup{X}$
is consistent, since it suffices to assign to the other $x_i$
variables the values dictated by $\mbf{v}$.
Furthermore, it is also plain that $R\cup{X}\entails{M}$ corresponds
to the set $X$ being sufficient for the prediction.
Therefore, abductive explanations are indeed an instantiation of
logic-based abduction, when explanations (for the logic-based
abduction problem) are restricted to being subset-minimal.

The use of abduction as outlined above is also exploited in practice
when using logic encodings for either computing AXps or CXps.
The set $S$ can be viewed as a set of soft clauses such that
$T\cup\{\neg{o}\}\cup{S}\entails\bot$. (This is equivalent to
$T\cup{S}\entails{M}$, with the single clause in $M$ complemented.)
The goal is then to find a minimal subset of $S$ of $H$, such that
inconsistency is preserved. This corresponds to finding a minimal
unsatisfiable subset (MUS) of an inconsistent propositional
formula~\cite{sat-handbook21}.
\end{document}